\pdfoutput=1

\documentclass[11pt]{article}

\usepackage[]{acl}
\usepackage{times}
\usepackage{latexsym}
\usepackage{graphicx, subfigure}
\usepackage{hyperref}

\usepackage[T1]{fontenc}

\usepackage[utf8]{inputenc}

\usepackage{microtype}

\title{SafeWebUH at SemEval-2023 Task 11: Learning Annotator Disagreement in Derogatory Text: Comparison of Direct Training vs Aggregation}

\author{Sadat Shahriar \\
  University of Houston\\
  \texttt{sshahriar@uh.edu} \\\And
  Thamar Solorio \\
  University of Houston \\
  \texttt{tsolorio@uh.edu} \\}

\begin{document}
\maketitle
\begin{abstract}
Subjectivity and difference of opinion are key social phenomena, and it is crucial to take these into account in the annotation and detection process of derogatory textual content. In this paper, we use four datasets provided by SemEval-2023 Task 11 and fine-tune a BERT model to capture the disagreement in the annotation. We find individual annotator modeling and aggregation lowers the Cross-Entropy score by an average of 0.21, compared to the direct training on the soft labels. Our findings further demonstrate that annotator metadata contributes to the average 0.029 reduction in the Cross-Entropy score.
\end{abstract}

\section{Introduction}

While the web space is inundated with derogatory textual content, the subjectivity of their interpretation frequently necessitates a system capable of capturing reader disagreements. The Learning-With-Disagreement (Le-Wi-Di) task involves learning annotators' disagreements based on how they categorize a text \cite{LeWiDi2023semeval}. Recent research has found that almost every annotation task contains a wide range of disagreements \cite{dumitrache-etal-2019-crowdsourced, pavlick-kwiatkowski-2019-inherent}. The subjective and biased nature of the raters, among other elements of natural language comprehension, make it crucial to learn disagreements through annotations ~\cite{uma-etal-2021-semeval}. In this study, we compare two strategies of disagreement learning: Disagreement Targeted Learning of soft labels, and annotator-specific learning with Post Aggregation, using BERT model. Furthermore, we utilize annotator-specific metadata, to capture annotators' disagreements in disparaging content.

Since the advent of social media, which has flooded the web with massive amounts of content, the number of offensive text, such as hate speech, misogyny, sexism, and abusive content has also increased significantly. Several studies were carried out to battle this problem, such as, Burnap and Williams studied online hate-speech in tweets, triggered by the murder of Lee Rigby, a London-based drummer~\cite{Burnap2015CyberHS}. Xu et al. formulated the cyber-bullying in social media as an NLP task~\cite{xu-etal-2012-learning}. Similar works are conducted in \citealt{warner-hirschberg-2012-detecting, silva2016analyzing, Gitari2015ALA}. However, tasks related to the detection of social phenomena, like offensiveness, and toxicity are often subjective in nature~\cite{9679002}. A recent survey among American adults stated that according to half of the participants, ``it is hard to know what others might find offensive'', and the majority of them acknowledged there were disagreements in what is perceived as sexist or racist~\cite{pewresearch}. To this end, we aim to develop a system that can capture subjective disagreement in derogatory text. 

The four datasets in the Le-Wi-di task come with the annotator-specific labels, with aggregated hard labels (majority voting) and soft labels (average of the labels). Although a system for modeling disagreements should be trained to estimate soft labels, it is not clear, whether direct training on the soft label or aggregating on the annotator labels is a better approach. Hence, our first research question \textbf{(Q1)}: Can annotator-specific classification models, and post hoc aggregation outperform the direct approach of regression on soft labels in disagreement modeling? Additionally, we explore the annotator metadata which explains how an annotator labeled other related text, and we pose the question \textbf{(Q2)}: Can annotator metadata improve the disagreement modeling? To address these questions, we compare BERT-based disagreement-targeted learning (regression) and post-aggregation learning (classification) and explore different strategies for incorporating annotator metadata to model the disagreement. However, due to the inconsistency of the annotators and lack of metadata, we limit our comparisons to two datasets only. 

Our work has several important implications. To begin, our model's ability to capture conflicts makes it applicable to the modeling of controversial social phenomena and public opinions. Hence, it can be used to model ambiguity in textual ambiguity. Furthermore, our explorations of incorporating annotator metadata can help understanding readers' perception and outlook in different context. Finally, enhancing transparency and accountability among the raters can be performed as a mean to quality control in multi-rater annotation process. The code for implementing our work is available here: \href{https://github.com/sadat1971/Le-Wi-Di-SemEval-23}{https://github.com/sadat1971/Le-Wi-Di-SemEval-23}

\begin{figure*}
    \centering
    \includegraphics[width=.90\textwidth]{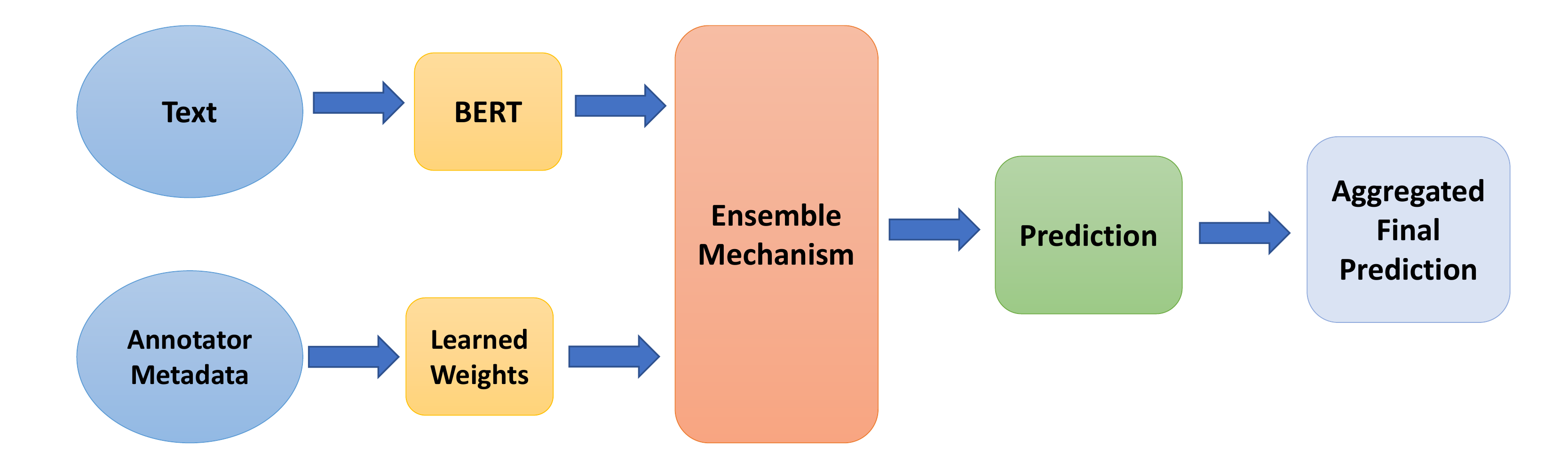}
    \caption{The text is fed to a pretrained BERT model, and fine-tuned for the downstream task. For the Post-Aggregation approach, the downstream classification task of the BERT model is to predict the label for each annotator. The softmax value from each annotator model with the metadata is ensembled to produce the annotator-specific soft-label and hard-label prediction.  For the Disagreement Targeted Learning approach, the BERT downstream task is to directly learn the soft labels. The ensemble mechanism is performed by regression approach to learn the final prediction.}
    \label{fig:model_arch}
\end{figure*}

\section{Dataset and Task Description}
SemEval'23 Task 11 has four datasets that deal with derogatory text. While the three datasets are in English, \textit{ArMIS} is in Arabic. Along with soft and hard labels, each dataset contains some metadata. They are described below in brief. 

The \textit{\textbf{MultiDomain Agreement}} (MD) dataset comes with tweets from three domains: BLM, Election and COVID-19 \cite{leonardelli-etal-2021-agreeing}. A total of 819 annotators were used to label all the tweets using AMT. A random combination of five annotators was chosen to label each tweet for offensiveness. The train set contains 6,592 tweets, the dev set from the practice phase has 1,104 tweets, and the test set from the evaluation phase contains 3,057 tweets.

The \textit{\textbf{HS-Brexit}} dataset contains tweets related to Brexit, and annotation from six annotators (a target group of three Muslim immigrants in the UK and a control group of three) \cite{akhtar2021whose}. Each of them labeled a tweet for hate speech, which is the target class of the task. They also annotated tweets for being offensive and aggressive. The train, dev, and test set have 784, 168, and 168 tweets respectively. 

Misogyny and Sexism are labeled in the \textit{\textbf{ArMIS}} dataset, rated by three annotators (Moderate Female, Liberal Female, and Conservative Male) \cite{almanea2022armis}. There are 657, 141, and 145 tweets in the train, dev, and test sets, respectively.

The \textit{\textbf{ConvAbuse}} dataset captures dialogues between a user and two conversational agents, and at least two annotators annotated the conversation for abusiveness \cite{cercas-curry-etal-2021-convabuse}. The dataset also provides labels for a conversation being sexist, explicit, implicit, intellectual, racist, transphobic, homophobic, and the target of the abuse. The train, dev, and test set have 2,398, 812, and 840 tweets respectively.

\section{System Description}

For the textual data, we use a pretrained language representation model, called BERT \cite{devlin-etal-2019-bert}. Since BERT is trained on the English data only, to handle the ArMIS task, we use Arabic-BERT \cite{safaya-etal-2020-kuisail}. Figure \ref{fig:model_arch} shows the system description. To address \textbf{Q1}, we compare two techniques-- Post Aggregation and Disagreement Targeted Learning, and we also investigate the effect of metadata to address \textbf{Q2}. The performance is measured by F1-score and Cross-Entropy (CE) score. 

\subsection{Post-Aggregation}
In the \textit{Post-Aggregation} (Post-Agg) approach, separate models are trained to learn the annotation pattern of each annotator. First, the BERT model is fine-tuned to learn the target class, and the softmax score $S$ is obtained for all annotators. Next, we process the metadata to extract important information. For the HS-Brexit dataset, in addition to labeling for hate speech, each annotator also labeled tweets for offensive and aggresive, which is available with the dataset. We compute the probability of a tweet being labeled as hate speech, given how it is labeled by an annotator as offensive and aggressive, which we denote as $P$. For each tweet, the soft label $\hat{SL}$ is then computed as, 
\begin{equation}
    \hat{SL}(w) = \frac{1}{N} \sum_{i=1}^{N} \frac{S_{i} + w*P_i}{1+w} 
\end{equation}

where N is the number of annotators. Since both $S_i$ and $P_i$ are predicted soft labels, we find their weighted average and select $w$, where the minimum CE score and maximum F1-score are obtained based on the dev set.

\subsection{Disagreement Targeted Learning}
While the Post-Agg approach considers learning from each annotator, the Disagreement Targeted Learning (Dis-Learning) approach learns only from the aggregated labels.
First, a BERT model is fine-tuned using a downstream regression task of estimating the soft label, and the predicted variable, $SL_{BERT}$ is obtained. Next, we measure the average rating of each metadata for all annotators across the entire dataset. For example, in HS-Brexit dataset, if two annotator labels a tweet as offensive, while four as not-offensive, the average metadata (offensiveness score) for that tweet will be $2/6 = 0.33$. Next, we train a linear regression model to predict the soft label based only on the available average metadata rating.

\begin{equation}
    SL_{meta} = b_{0} + b_{1}*M_{1} + b_{2}*M_{2}
\end{equation}
$b_{0}, b_{1}, b_{2}$ are trained from the linear regression model, and $M_{1}$ and $M_{2}$ are two metadata scores. For HS-Brexit, we use average offensive and aggressive measures. For the ConvAbuse dataset, out of twelve metadata labels, we pick the top two, \textit{explicit} and \textit{target system} which yielded the best correlation coefficient with the soft label values. Finally, we find $SL$ by averaging $SL_{BERT}$ and $SL_{meta}$.

\begin{table*}[t!]
\begin{center}
\begin{tabular}{l|cc|cc||cc|cc}

 \cline{2-9} &
 \multicolumn{2}{c}{\textbf{Post-Agg}} &  \multicolumn{2}{c||}{\textbf{Post-Agg-meta}} & \multicolumn{2}{c}{\textbf{Dis-Learning}} & \multicolumn{2}{c}{\textbf{Dis-Learning-meta}}\\
 \hline
\textbf{Dataset} & F1 & CE & F1 & CE & F1 & CE & F1 & CE\\
\hline
\hline
\textbf{MD} & -- & -- & -- & -- & \textbf{0.8266}* & \textbf{0.5076}* & -- & -- \\
\textbf{HS-Brexit} & 0.8810 & 0.1686 & \textbf{0.9167} & \textbf{0.0834} & 0.8869 & 0.3086 & 0.9107* & 0.2792* \\
\textbf{ArMIS} & 0.7211 & \textbf{0.2683} & -- & -- & \textbf{0.7586}* & 0.5753* & -- & -- \\
\textbf{ConvAbuse} & -- & -- & -- & -- & 0.9321* & 0.2364* & \textbf{0.9667} & \textbf{0.0688} \\
\hline
\end{tabular}
\end{center}
\caption{Comparing the performances of disagreement modeling approaches in all four datasets. Because the MD and ConvAbuse datasets lack consistent annotators, their Post-Agg results are not reported. Also, MD and ArMIS lack annotator-specific metadata, and thus, their metadata-incorporated performance is not reported as well. The best performance in each dataset is denoted in bold numbers, whereas the performance submitted in the Le-Wi-Di task is indicated with asterisks (*).}
\label{tab:result}
\end{table*}

\section{Experimental Set-up}
All of our models use ``bert-base-uncased'' version of BERT (``bert-base-arabic'' in ArMIS). We deploy a two-layered fully-connected network for fine-tuning in both regression and classification tasks. We choose the hyper-parameters from all the combinations, by three-fold cross-validation in the practice phase, and on the released validation set in the evaluation phase. The hidden size and dropout rate are chosen from \{32, 64, 128, 256\}, and \{.1, .3, .5\}. The learning rate is chosen from \{5e-4, 1e-5, 5e-5, 1e-6\}. We keep the batch size small due to the GPU limitations and choose from \{8, 16\}. Since BERT models quickly overfit on the data, we kept the epoch size between 2 and 4. However, for ``arabic-bert-base'', the performance was unstable, and we train upto 10 epochs. For all cases, AdamW is used as optimizer \cite{loshchilov2017decoupled}. For all our experiments, Pytorch version 1.11.0 is used \cite{NEURIPS2019_9015}.

To evaluate the result of capturing disagreement, we use the Cross-Entropy score provided by the competition. If the target soft label is $T$, and predicted soft label is $P$, for a dataset of size $D$, the Cross-Entropy (CE) is computed as:
\begin{equation}
    CE = -\frac{1}{D}\sum_{i=1}^{D} T_{i}*log(P_{i} + 1e-9) 
\end{equation}

We further report the F1-score (micro) on the hard label to evaluate the model performance on the majority-voted final prediction task. 

\section{Result and Discussion}
Table \ref{tab:result} shows that for HS-Brexit dataset, the Post-Agg approach does not improve the F1-score from the Dis-Learning approach. However, the Post-Agg approach is able to reduce the CE score by 0.1400 from the Dis-Learning approach. The reduction is even higher when metadata is used (by 0.1958). Similarly, for the ArMIS dataset, Dis-Learning approach has higher F1-score compared to the Post-Agg approach, while the CE score is lower in the Dis-Learning approach (reduced by 0.3070).  

We further investigate why the Post-Agg approach works better at capturing disagreement. Since the Dis-Learning approach does not take into account individual annotators, it mainly approximates the ``intensity'' of a text being derogatory. Conversely, the Post-Agg approach considers each annotator separately and learns their annotation pattern, which is aggregated afterward. Consequently, Dis-Learning has to depend only on textual data, making its job harder than Post-Agg. However, in a realistic case, the annotators may not be consistent (as in the MD and ConvAbuse datasets), or a large number of models are needed to be trained, rendering the Post-Agg technique infeasible.  Therefore, the Post-Agg approach is better suited for modeling disagreement if a small number of annotators are consistent across the dataset. Hence, \textbf{Q1} is addressed.

Next, the results reveal that performance is enhanced when annotator metadata is utilized as opposed to when it is not (Table \ref{tab:result}). Using the metadata reduced the CE score for the HS-Brexit dataset by 0.0852 and 0.0294 for the Post-Agg and Dis-Learning approaches, respectively. Similarly, for the ConvAbuse dataset, annotator metadata helps lower the CE score by 0.1676. The metadata contains useful annotation patterns of the annotators, which ameliorates the learning process. Notably, we have not used the metadata from MD and ArMIS, since they do not contain the related annotation information from the annotators. Therefore, \textbf{Q2} is addressed. 

In the MD, HS-Brexit, ArMIS, and ConvAbuse datasets, our results were ranked 7th, 9th, 11th, and 12th, respectively. Overall, we ranked 9th in the CE score category and 8th in the F1-score category.

\paragraph{Error Analysis}

Finally, we focus on the error analysis of this study. We find that both our approaches often make mistakes in prediction for the texts that do not use slang or curse words but are still voted by the majority as offensive. For example, three of the five annotators annotated the following sentence as offensive (soft label 0.60): \textit{\#TonyBobulinski  \#MAGA2020 \#MAGA \#ChangeYourVoteToTrump \#BidenCrimeFamily \#BidenHarris2020 \#BidenCares \#LaptopFromHell   Joe is going down. <url>}. However, our model predicts the soft label as 0.15. Similarly, tweets that contain curse words but do not necessarily exhibit offensiveness, are sometimes mistaken by our model as hate speech. For example, the tweet: \textit{Astounding Words from the prolific and talented - <user>  \#BlackLivesMatter \#fucktrump <url>} is labeled as non-offensive by three annotators out of five, however, our model predicts the soft label as 0.85, due to the presence of profane language in one of the hashtags.

\section{Related Works}
Though the majority of AI learning still operates under the assumption that a single interpretation exists for each item, research is growing to build learning methods that do not rely on this assumption \cite{uma2021learning}. Rater's Disagreement is a familiar phenomenon in Natural Language Processing \cite{poesio-artstein-2005-reliability, RECASENS20111138}. The disagreement may take place because of the annotator error or interface problem \cite{plank-etal-2014-linguistically}, explicit or implicit ambiguity \cite{poesio-artstein-2005-reliability}, item difficulty \cite{zaenen2005local}, and subjectivity \cite{akhtar2019new}. Notwithstanding, the simpler task such as POS tagging \cite{plank-etal-2014-linguistically} to subjective tasks like sentiment analysis, semantic role assignments also involve raters' disagreement \cite{kenyon-dean-etal-2018-sentiment, dumitrache-etal-2019-crowdsourced}.  Hence, researchers argued for taking disagreement into account during the labeling process and retaining the implicit ambiguity \cite{recasens-etal-2012-annotating, poesio-artstein-2005-reliability}). To this end, we explore the Learning-With Disagreement task for derogatory text. 

The previous version of this competition was launched in 2021, where the organizers used NL and image-classification task to address for disagreement in the labeling \cite{uma-etal-2021-semeval}. The winning team used the Sharpness-Aware Minimization technique (SAM) and a special NN layer called softmax Crowd-layer with BERT as baseline model \cite{osei-brefo-etal-2021-uor}. While the SAM architecture was mainly used for CIFAR-10 (image classification), the Crowdlayer architecture aims to map the label with each individual annotator. Since the current competition only involves text, we fine-tune a BERT model and use the annotator metadata to capture the disagreement. 

\section{Conclusion}  

Because of the proliferation of social media content, the internet has become a breeding ground for derogatory text. However, due to the differences in human perception and opinion, often there is no unanimous consensus among the annotators about the text being derogatory or not. Hence, it is imperative to store the soft labels and capture annotator disagreement in the modeling process. Our work compares the direct training on the soft label with the annotator-specific model and post-aggregation. We find that with the presence of consistent annotators, it might be helpful to take the latter approach. In addition, integrating annotator metadata has been proved to be beneficial in our experiments. Our work has a wide variety of potential future research directions, such as:

\begin{itemize}
    \item We only modeled with one Transformer-based approach, BERT. In the future, we plan to use RoBERTA, ELECTRA and XLMNet
    \item We find a strong correlation between hate speech and offensiveness. Therefore, we plan to investigate how cross-dataset performance works. Such experiments will also help to make our model more generalizable.
    \item Because language evolves in response to social context and other phenomena, it is critical to include Continual Learning (CL) techniques and investigate the distribution shift in the annotation process. In the future, we intend to incorporate CL into our work.
\end{itemize}

\bibliography{anthology,custom}



\end{document}